\pdfoutput=1

\documentclass[11pt]{article}
\usepackage{authblk}

\usepackage{naacl2021}

\usepackage{times}
\usepackage{latexsym}

\usepackage[T1]{fontenc}

\usepackage[utf8]{inputenc}

\usepackage{microtype}

\usepackage{times}
\usepackage{latexsym}
\usepackage{amsmath, amsfonts}
\usepackage{url}
\usepackage{mathdots}
\usepackage{yhmath}
\usepackage{cancel}
\usepackage{color}
\usepackage{siunitx}
\usepackage{array}
\usepackage{multirow}
\usepackage{amssymb}
\usepackage{graphicx}
\usepackage{float}
\usepackage{rotating}
\usepackage{tabularx}
\usepackage{booktabs}
\usepackage{svg}
\usepackage{tablefootnote}
\usepackage{cleveref}
\usepackage{tabularx}
\usepackage{xcolor}
\usepackage{makecell}
\usepackage{enumitem}
\usepackage{multirow}
\usepackage{haldefs}
\usepackage{todonotes}
\usepackage{subcaption}

\renewcommand{\sectionautorefname}{\S\kern-0.2em}
\renewcommand{\subsectionautorefname}{\S\kern-0.2em}
\renewcommand{\subsubsectionautorefname}{\S\kern-0.2em}

\newcommand{\newpara}[1]{\noindent \textbf{#1} \hspace{0.5em}}

\newcommand{\naacl}[1]{#1}

\makeatletter
\renewcommand*{\@fnsymbol}[1]{\ensuremath{\ifcase#1\or \dagger\or \ddagger\or
    \mathsection\or \mathparagraph\or \|\or **\or \dagger\dagger
    \or \ddagger\ddagger \else\@ctrerr\fi}}
\makeatother

\title{Ask what's \emph{missing} and what's \emph{useful}: \\Improving Clarification Question Generation using Global Knowledge}

\author[$\clubsuit$]{\textbf{Bodhisattwa Prasad Majumder\thanks{\hspace{0.5em}Work done during an internship at Microsoft Research}\protect\phantom{\footnotesize 1}}}
\author[$\diamondsuit$]{\textbf{Sudha Rao}}
\author[$\diamondsuit$]{\textbf{\quad\quad\quad\quad\quad\quad\quad\quad\quad\quad\quad\quad\quad\quad Michel Galley}}
\author[$\clubsuit$]{\textbf{Julian McAuley}}
\affil[$\clubsuit$]{Department of Computer Science and Engineering, UC San Diego \protect\\ \tt \{bmajumde, jmcauley\}@eng.ucsd.edu}
\affil[$\diamondsuit$]{Microsoft Research, Redmond \protect\\ \tt \{sudha.rao, mgalley\}@microsoft.com}

\makeatletter
\renewcommand\outauthor{
    \begin{tabular}[t]{>{\centering}p{14cm}} 
    \bf\@author
    \end{tabular}}
\makeatother

\begin{document}
\maketitle
\begin{abstract}

The ability to generate clarification questions i.e.,~questions that identify useful missing information in a given context, is important in reducing ambiguity. 
Humans use previous experience with similar contexts to form a global view and compare it to the given context to ascertain what is missing and what is useful in the context. 
Inspired by this, we propose a model for clarification question generation where we first identify what is missing by taking a  difference between the global and the local view and then train a model to identify what is useful and generate a question about it. Our model outperforms several baselines as judged by both automatic metrics and humans.

\end{abstract}

\section{Introduction}

\begin{figure*}
\includegraphics[trim=0 140 0 110, width=0.95\linewidth]{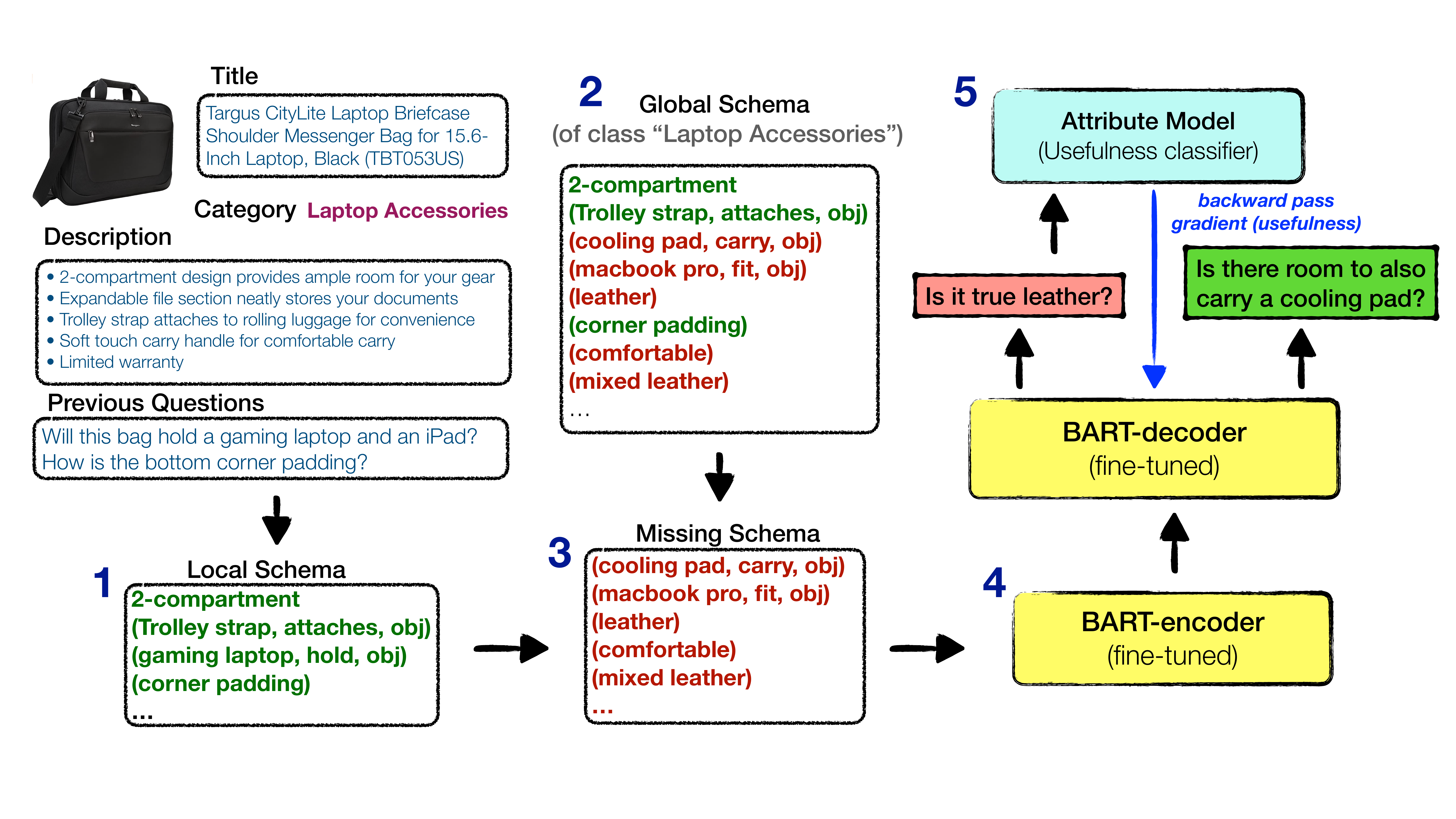}
\centering
\caption{Test-time behaviour of our proposed model for \emph{useful} clarification question generation based on \emph{missing} information in a Community-QA (amazon.com) setup. 1. We obtain a local schema from the available context for a product: description and previously asked questions. 2. We obtain the global schema of the category of the product. 3. We estimate the \emph{missing} schema that is likely to guide clarification question generation. 4. A BART model fine-tuned on (missing schema, question) pairs to generate a question (``\textit{Is it true leather?}''). 5. A PPLM model with usefulness classifier as its attribute model further tunes the generated question to make it more \emph{useful} (``\textit{Is there room to also carry a cooling pad?}''). 
}
\label{fig:model-diagram}
\vspace{-1em}
\end{figure*}

An important but under-explored aspect of text understanding is the identification of \emph{missing information in a given context} i.e.,~information that is essential to accomplish an underlying goal but is currently missing from the text. 
Identifying such missing information can help to reduce ambiguity in a 
given context which can aid machine learning models in prediction and generation \cite{de2003analysis,stoyanchev2014towards}.
\citet{rao2018learning, rao-daume-iii-2019-answer} recently proposed the task of clarification question generation as a way to identify such missing information in context. They propose a model for this task which while successful at generating fluent and relevant questions, still falls short in terms of usefulness and identifying missing information.
With the advent of large-scale pretrained generative models \cite{radford2019language,lewis2019bart,raffel2019exploring}, generating fluent and coherent text is within reach. However, generating clarification questions requires going beyond fluency and relevance.
Doing so requires understanding what is missing, which if included could be useful to the consumer of the information. 

Humans are naturally good at identifying missing information in a given context. They possibly make use of \emph{global knowledge} i.e.,~recollecting previous similar contexts and comparing them
to the current one to ascertain what information is \textit{missing} and if added would be the most \textit{useful}. 
Inspired by this, we propose a two-stage framework for the task of clarification question generation. Our model hinges on the concept of a ``schema'' which we define as the key pieces of information in a text.
In the first stage, we find \textit{what's missing} by taking a difference between the global knowledge's schema and schema of the local context (\autoref{sec:identifying-missing-info}). In the second stage we feed this missing schema to a fine-tuned BART \cite{lewis2019bart} model to generate a question which is further made more \textit{useful} using PPLM \cite{dathathri2019plug} (\autoref{sec:identify-usefulness}).\footnote{The code is available at \url{https://github.com/microsoft/clarification-qgen-globalinfo}} 


\begin{table}[t]
\centering
\footnotesize
\begin{tabular}{@{}rp{5.8cm}@{}}
\toprule
  \textsc{Title:} & Sony 18x Optical Zoom 330x Digital Zoom Hi8 Camcorder \\[0.5em]
  \textsc{Desc:} & Sony Hi-8mm Handycam Vision camcorder 330X digital zoom, Nightshot(TM) Infrared 0 lux system, Special Effects, 2.5" SwivelScreen color LCD and 16:9 recording mode, Laserlink connection. Image Stabilization, remote, built in video light.\\[0.5em]
\textsc{Question:} & Can I manually control the video quality?\\[0.5em]
\bottomrule
\end{tabular}
\caption{Product description from amazon.com paired with a clarification question generated by our  model.}\label{amazon-eg}
\vspace{-1.5em}
\end{table}

We test our proposed model on two scenarios (\autoref{sec:setup-scenarios}): \emph{community-QA}, where the context is a product description from amazon.com \cite{mcauley2016addressing} (see e.g.~\autoref{amazon-eg});
and \emph{dialog} where the context is a dialog history
from the
Ubuntu Chat forum \cite{lowe2015ubuntu}. 
We compare our model to several baselines (\autoref{sec:baselines}) and evaluate outputs using both automatic metrics and human evaluation to show that our model significantly outperforms baselines in generating useful questions that identify missing information in a given context (\autoref{sec:experimental-results}).
Furthermore, our analysis reveals reasoning behind generated questions as well as robustness of our model to available contextual information. 
(\autoref{sec:analysis}). 


\section{Problem Setup and Scenarios}\label{sec:setup-scenarios}

\citet{rao2018learning} define the task of clarification question generation as: given a context, generate a question that identifies missing information in the context. We consider two scenarios:

\paragraph{Community-QA}
Community-driven question-answering has become a common venue for crowdsourcing answers. These forums often have some initial context on which people ask clarification questions. We consider the Amazon question-answer dataset \cite{mcauley2016addressing} where context is a product description and the task is to generate a clarification question that helps a potential buyer better understand the product. 



\paragraph{Goal Oriented Dialog}
With the advent of high quality speech recognition and text generation systems, we are increasingly using dialog as a mode to interact with devices \cite{DBLP:conf/chi/ClarkPCDGESGMMW19}. However, these dialog systems still struggle when faced with ambiguity and could greatly benefit from having the ability to ask clarification questions. 
We explore such a goal-oriented dialog scenario using the Ubuntu Dialog Corpus \cite{lowe2015ubuntu} consisting of dialogs between a person facing a technical issue and another person helping them resolve the issue. Given a context i.e a dialog history, the task is to generate a clarification question that would aid the resolution of the technical issue.


\section{Approach}\label{sec:proposed-model}

\autoref{fig:model-diagram} depicts our approach at a high level.
We propose a two-stage approach for the task of clarification question generation. In the first stage, we identify the missing information in a given context. For this, we first group together all similar contexts in our data\footnote{See \autoref{sec:dataset} for details to combine data splits} to form the \textit{global schema} for each high-level class. Next, we extract the schema of the given context to form the \textit{local schema}. Finally, we take a difference between the local schema and the global schema (of the class to which the context belongs) to identify the missing schema for the given context. In the second stage, we train a model to generate a question about the most useful information in the missing schema. For this, we fine-tune a BART model \cite{lewis2019bart} on (missing schema, question) pairs and at test time, we use PPLM \cite{dathathri2019plug} with a usefulness classifier as the attribute model to generate a useful question about missing information. 

\subsection{Identifying Missing Information}\label{sec:identifying-missing-info}

\newpara{Schema Definition}
Motivated by \cite{khashabi2017learning} who use essential terms from a question to improve performance of a Question-Answering system, we see the need of identifying important elements in a context to ask a better question. 
We define schema of sentence $s$ as set consisting of one or more triples of the form (key-phrase, verb, relation) and/or one or more key-phrases. 
\begin{equation}
\begin{split}
        \mathit{schema}_{s} & = \{ \: \mathit{element} \:\}; \: \text{where}  \\
        \mathit{element} \in \{ &  (\mathit{key\mbox{-}phrase}, \mathit{verb}, \mathit{relation}), \\
        & \mathit{key\mbox{-}phrase} \} 
\end{split}
\end{equation}

\begin{figure}[h!]
\includegraphics[trim=0 0 0 0, width=\linewidth]{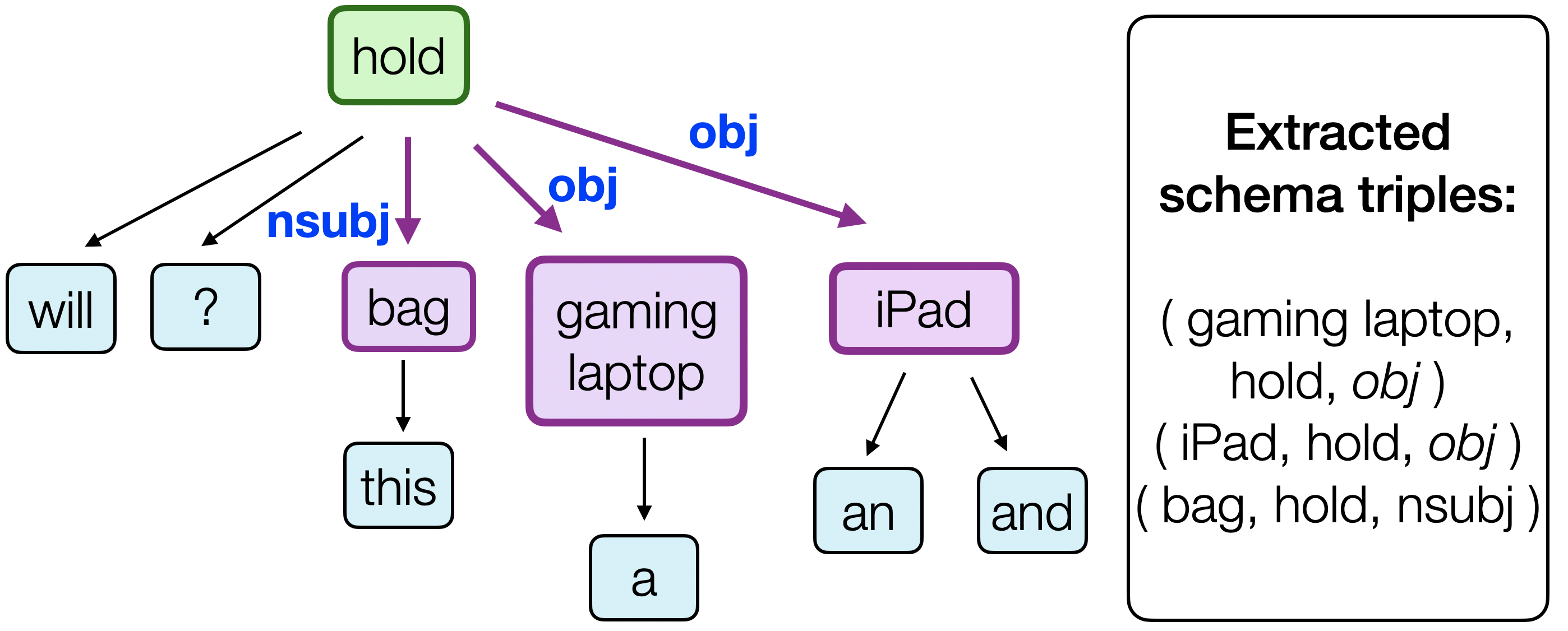}
\centering
\caption{Dependency tree and paths showing how we obtain schema triples
for a sentence: \emph{``Will this bag hold a gaming laptop and an iPad?''} (from \autoref{fig:model-diagram}).
}
\label{fig:dep-tree}
\end{figure}

\newpara{Schema Extraction}
Our goal is to extract a schema from a given context. 
We consider (key-phrase, action verb, relation) as the basic element of our schema. Such triples have been found to be representative of key information in previous work \cite{DBLP:journals/corr/abs-1904-08524}. 
Given a sentence from the context, we first extract bigram and unigram key-phrases using YAKE (Yet-Another-Keyword-Extractor) \cite{campos2020yake} and retain only those that contain at
least a noun. 
We then obtain the dependency parse tree \cite{DBLP:conf/acl/QiZZBM20} of the sentence and map the key-phrases to tree nodes.\footnote{In the case of bigram phrases, we merge the tree nodes.}
Now, to obtain the required triple, we need to associate a verb and a relation to each key-phrase. 
This procedure is described in 
\autoref{alg:path-finding}.
At a high-level, we use the path between the key-phrase and the closest verb in the dependency tree to establish a relation between the key-phrase and the verb. In cases where there is no path, we use only the key-phrase as our schema element. 
\autoref{fig:dep-tree} shows an example dependency tree for a sentence.

\begin{algorithm}[t!]
\small
\begin{algorithmic}
\STATE Initialize with empty path (path length $\infty$) for all possible pairs of verbs ($\in$ \{VB, VBG, VBZ\}) and key-phrases in the sentence
\FOR{Each verb and key-phrase pair}
\STATE Search 
for the
key-phrase among the children of the verb in the dependency tree
\IF{A key-phrase is found \AND path is shorter than the stored path}
\STATE Update the path between the key-phrase and the verb pair
\ENDIF
\ENDFOR
\FOR{Each verb and key-phrase pair}
\IF{The key-phrase is the immediate child of the verb}
\STATE Create the triple (key-phrase, verb, relation) using the relation in the path
\ELSE
\STATE Traverse backward from the key-phrase, stop at the first verb, use the relation with its immediate child in the path to create (key-phrase, verb, relation)
\ENDIF
\ENDFOR
\end{algorithmic}
\caption{\small Pseudocode for extracting (key-phrase, verb, relation) triple.}
\label{alg:path-finding}
\end{algorithm}

\newpara{Creating local schema}
Given a context, we extract a schema for each sentence in the context. The local schema of a context $c$ is a union of schemata of each sentence $s$ in the context. 
\begin{equation}
    \mathit{local\_schema}_c = \cup_{s \in c} \: \mathit{schema}_s
\end{equation}

\newpara{Creating global schema}
We define global schema at the class level where a `class' is a group of similar contexts. 
For \textit{Amazon}, classes consist of groups of similar products and for \textit{Ubuntu}, classes consist of groups of similar dialogs (see \autoref{sec:dataset} for details). 
The global schema of a class $K$ is a union of local schemata of all contexts $c$ belonging to $K$. 
\begin{equation}
    \mathit{global\_schema}_{K} = \cup_{c \in K} \: \mathit{local\_schema}_c
\end{equation}

A naive union of all local schemata can result in a global schema that has a long tail of low-frequency schema elements. Moreover, it may have redundancy where schema elements with similar meaning are expressed differently (e.g.~\textit{OS} and \textit{operating system}). We therefore use word embedding based similarity to group together similar key-phrases and retain only the most frequent elements (see appendix).

\newpara{Creating a missing schema}
Given a context $c$, we first determine the class $K$ to which the context belongs. We then compute its missing schema by taking the set difference between the global schema of class $K$ and the local schema of the context $c$:
\begin{equation}
    \mathit{missing\_schema}_c = \mathit{global}_{K} \setminus \mathit{local}_c
\end{equation}

More specifically, we start with the elements in the global schema and remove elements that have a semantic match (see appendix)
with any element in the local schema to obtain the missing schema.

\subsection{Generating Useful Questions}\label{sec:identify-usefulness}

Our goal is to generate a useful question about missing information. In \autoref{sec:identifying-missing-info}, we explained how we compute the missing schema for a given context; here we describe how we train a model to generate a useful question given the missing schema. 

\paragraph{BART-based generation model}
Our generation model is based on the BART \cite{lewis2019bart} encoder-decoder model, which is also a state-of-the-art model in various generation tasks including dialog generation and summarization. We start with the pretrained base BART model consisting of a six layer encoder and six layer decoder. We fine-tune this model on our data where the inputs are the missing schema and 
the output is the question. The elements of
the
missing schema in the input are separated by a special \texttt{[SEP]} token. Since the elements in our input do not have any order, we use the same positional encoding for all input positions. We use a token type embedding layer with three types of tokens: key-phrases, verbs, and relations.

\paragraph{PPLM-based decoder} 
We observed during our human evaluation\footnote{See results of BART+missinfo in  \autoref{tab:amazon-human-eval-table}} that a BART model fine-tuned in this manner, in spite of generating questions that ask about missing information, does not always generate \emph{useful} questions. We therefore propose to integrate the usefulness criteria into our generation model. We use the Plug-and-Play-Language-Model (PPLM) \cite{dathathri2019plug} during decoding (at test time). The attribute model of the PPLM in our case is a usefulness classifier trained on bags-of-words of questions. In order to train such a classifier, we need usefulness annotations on a set of questions. For the Amazon dataset, we collect usefulness scores (0 or 1) on 5000 questions using human annotation whereas for the Ubuntu dataset we assume positive labels for (true context, question) pairs and negative labels for (random context, question) pairs and use 5000 such pairs to train the usefulness classifier. Details of negative sampling for Ubuntu dataset is in Appendix.

\section{Experiments}

We aim to answer the following research questions (RQ):
\begin{enumerate}[noitemsep,nolistsep]
\item Is the model that uses missing schema better at identifying missing information compared to models that use the context directly to generate questions?
\item Do large-scale pretrained models help generate better questions?
\item Does the PPLM-based decoder help increase the usefulness of the generated questions?
\end{enumerate}

\subsection{Datasets}\label{sec:dataset}

\newpara{Amazon} 
The Amazon review dataset \cite{mcauley2015image} consists of descriptions of products on amazon.com and the Amazon question-answering dataset \cite{mcauley2016addressing} consists of questions (and answers) asked about products. Given a product description and $N$ questions asked about the product, we create $N$ instances of (\textit{context}, \textit{question}) pairs where \textit{context} consists of the description and previously asked questions (if any). 
We use the ``Electronics'' category consisting of 23,686 products.
We split this into train, validation and test sets (\autoref{tab:data-statistics}).
The references for each context are all the questions (average=6) asked about the product. A class is defined as a group of products within a subcategory (e.g.~DSLR Camera) as defined in the dataset. We restrict a class to have at most 400 products, and a bigger subcategory is broken into lower-level subcategories (based on the product hierarchy) resulting in 203 classes. 
\naacl{While creating global schema, we exclude target questions from validation and test examples. The product descriptions and associated metadata come as inputs during test time. Hence, including them from all splits while creating the global schema does not expose the test and validation targets to the model during training.}

\begin{table}
    \centering
    \footnotesize
    \begin{tabular}{l | c c c}
              & Train & Validation & Test \\
    \hline
    \hline
      Amazon  & 123,567 &  4,525 &  2,361 \\
      Ubuntu  & 102,678 &  7,864 &  200 \\
    \end{tabular}
    \caption{Number of data instances in the train, validation and test splits of Amazon and Ubuntu datasets (Both datasets are in English. Links are in appendix)}
    \label{tab:data-statistics}
    \vspace{-1em}
\end{table}

\newpara{Ubuntu}
The Ubuntu dialog corpus \cite{lowe2015ubuntu} consists
of
utterances of dialog between two users on the Ubuntu chat forum. Given a dialog, we identify utterances that end with a question mark. We then create 
data instances of (context, question) where the question is the utterance ending with a question mark and the context consists of all utterances before the question. We consider only those contexts that have at least five utterances and at most ten utterances. \autoref{tab:data-statistics} shows the number of data instances in the train, validation and test splits. 
Unlike the Amazon dataset, each context has only one reference question. 
A class is defined as a group of dialogs that address similar topics. Since such class information is not present in the dataset, we use $k$-means to cluster dialogs into subsequent classes. Each dialog was represented using a TF-IDF vector. After tuning the number of clusters based on sum of squared distances of dialogs to their closest cluster center, we obtain 26 classes.
\naacl{We follow a similar scheme as with Amazon for not including target questions from validation and test sets while building the global schema.}

\subsection{Baselines and Ablations}\label{sec:baselines}

\newpara{Retrieval} We retrieve the question from the train set whose schema overlaps most with the missing schema of the given context.

\newpara{GAN-Utility} The state-of-the-art model for the task of clarification question generation \cite{rao-daume-iii-2019-answer} trained on (context, question, answer) triples. 

\newpara{Transformer} 
A transformer \cite{vaswani2017attention}\footnote{We use original hyperparameters \& tokenization scheme.} model trained on (context, question) pairs.

\newpara{BART} 
We finetune a BART model \cite{lewis2019bart} on (context, question) pairs.  

\newpara{BART + missinfo} 
We compare to a BART model fine-tuned on (missing schema, question) pairs. 

\newpara{BART + missinfo + WD} 
This is similar to the ``BART + missinfo'' baseline with the modification that, at test time only, we use a weighted-decoding (WD) strategy \cite{ghazvininejad2017hafez} by redefining the probability of words in the vocabulary using usefulness criteria (more in appendix).

\newpara{BART + missinfo + PPLM} This is our proposed model as described in \autoref{sec:proposed-model} where we fine-tune the BART model on (missing schema, question) pairs and use a usefulness classifier based PPLM model for decoding at test time. 

\subsection{Evaluation Metrics}

\begin{table*}[t!]
\centering
\footnotesize
\begin{minipage}{0.48\textwidth}
\centering
\begin{tabular}
{l c c c c }
Model &  \hspace{-1.5mm}BLEU-4 & \hspace{-1.5mm}METEOR & \hspace{-1.5mm}Distinct-2  \\
\hline
\hline
Retrieval & \hspace{-2mm}8.76 & \hspace{-2mm}9.23 &  \hspace{-2mm}\textbf{0.92} \\
GAN-Utility & \hspace{-2mm}14.23 & \hspace{-2mm}16.82 & \hspace{-2mm}0.79\\
Transformer & \hspace{-2mm}12.89 & \hspace{-2mm}14.56 & \hspace{-2mm}0.60 \\
BART & \hspace{-2mm}15.98 & \hspace{-2mm}16.78 & \hspace{-2mm}0.78 \\
\hspace{0.6em} + missinfo & \hspace{-2mm}16.87 & \hspace{-2mm}17.11 & \hspace{-2mm}0.82 \\
\hspace{0.6em} + missinfo + WD & \hspace{-2mm}16.23 &\hspace{-2mm}\textbf{ 17.98} & \hspace{-2mm}\textbf{0.84} \\
\hspace{0.6em} + missinfo + PPLM & \hspace{-2mm}\textbf{18.55} &  \hspace{-2mm}\textbf{18.01} & \hspace{-2mm}\textbf{0.86} \\
Reference  & \hspace{-2mm}-- & \hspace{-2mm}-- & \hspace{-2mm}0.95 \\
\hline
\end{tabular}
\caption{\label{tab:amazon-auto-eval-table} Automatic metric results on the full test set of Amazon. The difference between bold and non-bold numbers is statistically significant with $p < 0.001$.}
\end{minipage}%
\hfill
\begin{minipage}{0.48\textwidth}
\centering
\begin{tabular}
{l c c c c }
Model &  \hspace{-1.5mm}BLEU-4 & \hspace{-1.5mm}METEOR & \hspace{-1.5mm}Distinct-2  \\
\hline
\hline
Retrieval & \hspace{-2mm}4.89 & \hspace{-2mm}5.12 & \hspace{-2mm}\textbf{0.82} \\
Transformer & \hspace{-2mm}6.89 & \hspace{-2mm}7.45 & \hspace{-2mm}0.67 \\
BART & \hspace{-2mm}8.23 & \hspace{-2mm}9.67 & \hspace{-2mm}0.72 \\
\hspace{0.6em} + missinfo & \hspace{-2mm}\textbf{9.54} & \hspace{-2mm}10.78 & \hspace{-2mm}\textbf{0.75} \\
\hspace{0.6em} + missinfo + PPLM & \hspace{-2mm}\textbf{10.02} & \hspace{-2mm}\textbf{11.65} &\hspace{-2mm}\textbf{ 0.79} \\
Reference & \hspace{-2mm}-- & \hspace{-2mm}-- & \hspace{-2mm}0.87 \\
\hline
\end{tabular}
\caption{\label{tab:ubuntu-auto-eval-table} Automatic metric results the full test set of Ubuntu. The difference between bold and non-bold numbers is statistically significant with $p < 0.001$.}
\end{minipage}
\vspace{-1em}
\end{table*}

\subsubsection{Automatic Metrics} 

\textbf{BLEU-4} \cite{papineni2002bleu} evaluates 4-gram precision between model generation and references. 
at the corpus level; \textbf{METEOR} \cite{banerjee2005meteor} additionally uses stem and synonym matches for similarity; and \textbf{Distinct-2} \cite{li2016diversity} measures diversity by calculating the number of distinct bigrams in model generations scaled by the total number of generated tokens. 

\subsubsection{Human Judgment}
Similar to \citet{rao-daume-iii-2019-answer}, we conduct a human evaluation on Amazon Mechanical Turk
to evaluate model generation on the four criteria below. Each generated output is shown with the context and is evaluated by three annotators.

\newpara{Relevance} We ask ``\textit{Is the question relevant to the context?}'' and let annotators choose between Yes (1) and No (0). 

\newpara{Fluency} We ask ``\textit{Is the question grammatically well-formed i.e. a fluent English sentence?}'' and let annotators choose between Yes (1) and No (0).

\newpara{Missing Information} We ask ``\textit{Does the question ask for new information currently not included in the context?}'' and let annotators choose between Yes (1) and No (0).

\newpara{Usefulness} We perform a comparative study where we show annotators two model-generated questions (in a random order) along with the context. 
For
Amazon, we ask ``\textit{Choose which of the two questions is 
more useful to a potential buyer of the product}''. 
For
Ubuntu, we ask ``\textit{Choose which of the two questions is more useful to the other person in the dialog}''.

\subsection{Experimental Results}\label{sec:experimental-results}

\subsubsection{Automatic Metric Results}

\newpara{Amazon} \autoref{tab:amazon-auto-eval-table} shows automatic metric results on Amazon. Under BLEU-4 and METEOR, the retrieval model performs the worst suggesting that picking a random question that matches the most with the missing schema does not always yield a good question. This strengthens the need of the second stage of our proposed model i.e.~%
BART + PPLM based learning. GAN-Utility, which is state-of-the-art on Amazon, outperforms the Transformer baseline suggesting that training a larger model (in terms of 
the
number of parameters) does not always yield better questions. BART, on the other hand, outperforms GAN-Utility suggesting the benefit of large-scale pretraining (RQ2). BART+missinfo further outperforms BART showing the value in training on missing schemata instead of training directly on the context (RQ1). A variation of this model that uses weighted decoding performs marginally better on METEOR but slightly worse of BLEU-4. Our final proposed model i.e.,~BART+missinfo+PPLM performs the best among all baselines across both BLEU-4 and METEOR.

Under diversity (Distinct-2), the retrieval model produces the most diverse questions (as also observed by \citet{rao-daume-iii-2019-answer}) since it selects among human written questions which tend to be more diverse compared to model generated ones. Among other baselines, transformer interestingly has the lowest diversity whereas GAN-Utility and BART come very close to each other. Model ablations that use missing schema produce more diverse questions further strengthening the importance of training on missing schema. Our model i.e.,~BART+missinfo+PPLM, in spite of outperforming
all baselines (except retrieval), is still far from reference questions in terms of diversity, suggesting room for improvement. 

\paragraph{Ubuntu}
\autoref{tab:ubuntu-auto-eval-table} shows the results of automatic metrics on Ubuntu.\footnote{We do not experiment with the GAN-Utility model (since it requires ``answers'') and the BART+missinfo+WD model (since usefulness labels are not obtained from humans).}
The overall BLEU-4 and METEOR scores are much lower compared to Amazon since Ubuntu has only one reference per context. 
Under BLEU-4 and METEOR scores, similar to Amazon, we find that the retrieval baseline has the lowest scores. Transformer baseline outperforms the retrieval baseline but lags behind BART, again showing the
importance of large-scale pretraining. The difference between the BLEU-4 scores of BART+missinfo and our final proposed model is not significant but their METEOR score difference is significant suggesting that our model produces questions that may be lexically different from references but have more semantic overlap with the reference set. Under Distinct-2 scores, we find the same trend as in Amazon, with the
retrieval model being the most diverse and our final model outperforming all other baselines. 

\subsubsection{Human Judgement Results}

\begin{table*}[t!]
\centering
\footnotesize
\begin{minipage}{0.48\textwidth}
\centering
\begin{tabular}{l c c c }
Model & \hspace{-1.5mm}Relevancy & \hspace{-1.5mm}Fluency & \hspace{-1.5mm}MissInfo \\
\hline
\hline
GAN-Utility & \hspace{-2mm}0.9 & \hspace{-2mm}0.86 & \hspace{-2mm}0.81 \\ 
BART & \hspace{-2mm}0.94 & \hspace{-2mm}0.92 & \hspace{-2mm}0.77\\ 
\hspace{0.6em} + missinfo & \hspace{-2mm}0.97 & \hspace{-2mm}0.92 & \hspace{-2mm}0.87\\ 
\hspace{0.6em} + missinfo + PPLM & \hspace{-2mm}\textbf{0.99} & \hspace{-2mm}\textbf{0.93} & \hspace{-2mm}\textbf{0.89}\\ 
Reference & \hspace{-2mm}0.96 & \hspace{-2mm}0.83 & \hspace{-2mm}0.89\\ 
\hline
\end{tabular}
\caption{Human judgment results (0-1) on 300 randomly sampled descriptions from the Amazon test set}\label{tab:amazon-human-eval-table}
\end{minipage}%
\hfill
\begin{minipage}{0.48\textwidth}
\centering
\begin{tabular}{l c c c }
Model & \hspace{-1.5mm}Relevancy & \hspace{-1.5mm}Fluency & \hspace{-1.5mm}MissInfo \\
\hline
\hline
Transformer & \hspace{-2mm}0.74 & \hspace{-2mm}0.99 & \hspace{-2mm}0.99 \\ 
BART & \hspace{-2mm}0.69 & \hspace{-2mm}0.99 & \hspace{-2mm}0.96\\ 
\hspace{0.6em} + missinfo & \hspace{-2mm}0.81 & \hspace{-2mm}0.95 & \hspace{-2mm}0.98 \\ 
\hspace{0.6em} + missinfo + PPLM & \hspace{-2mm}0.91 & \hspace{-2mm}0.83 & \hspace{-2mm}0.99\\ 
Reference & \hspace{-2mm}0.85 & \hspace{-2mm}0.83 & \hspace{-2mm}0.96\\ 
\hline
\end{tabular}
\caption{\label{tab:ubuntu-human-eval-table} Human judgment results (0-1) on 150 randomly sampled dialog contexts from Ubuntu test set}
\end{minipage}
\end{table*}

\begin{figure*}[t!]
\begin{subfigure}{.48\linewidth}
  \centering
  \includegraphics[scale=0.48]{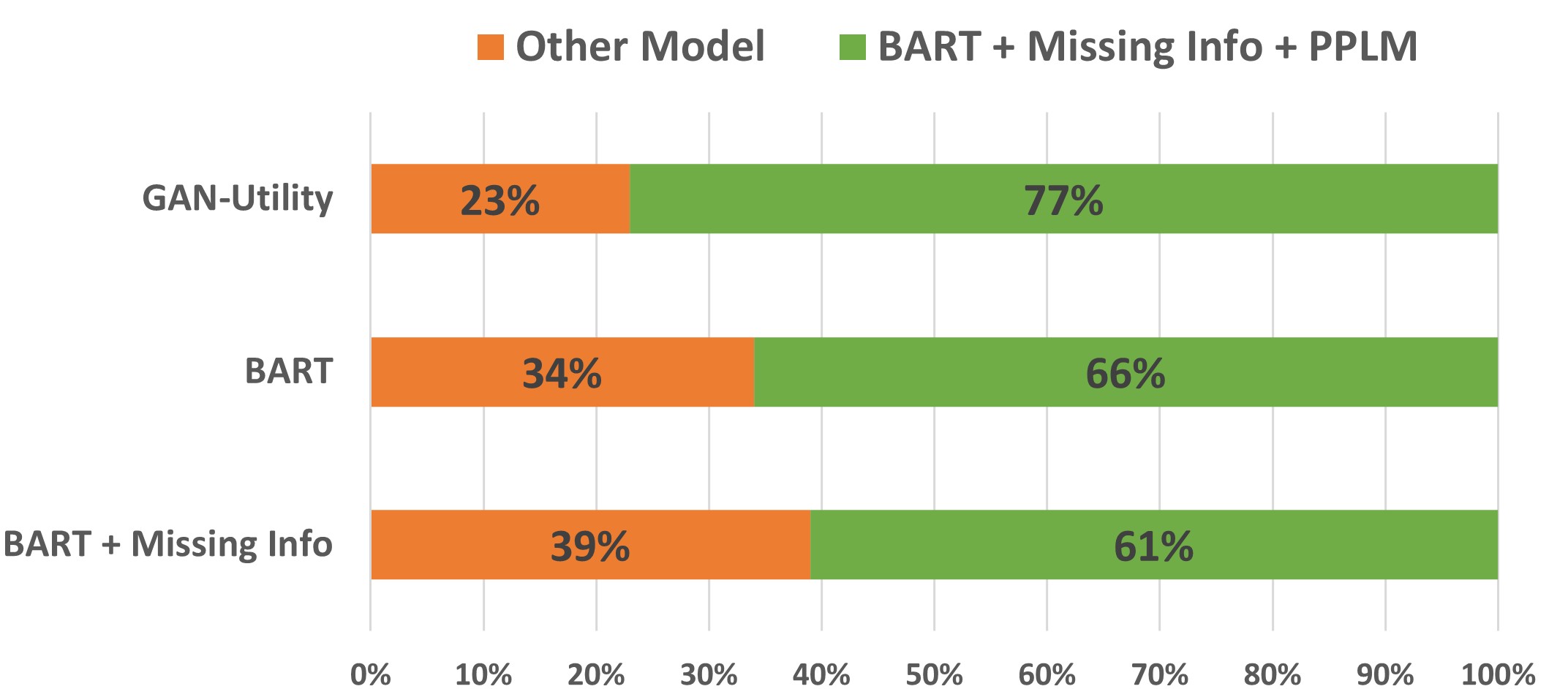}  
  \caption{}
  \label{fig:amazon-usefulness-compare-results}
\end{subfigure}
\begin{subfigure}{.48\linewidth}
  \centering
  \includegraphics[scale=0.48]{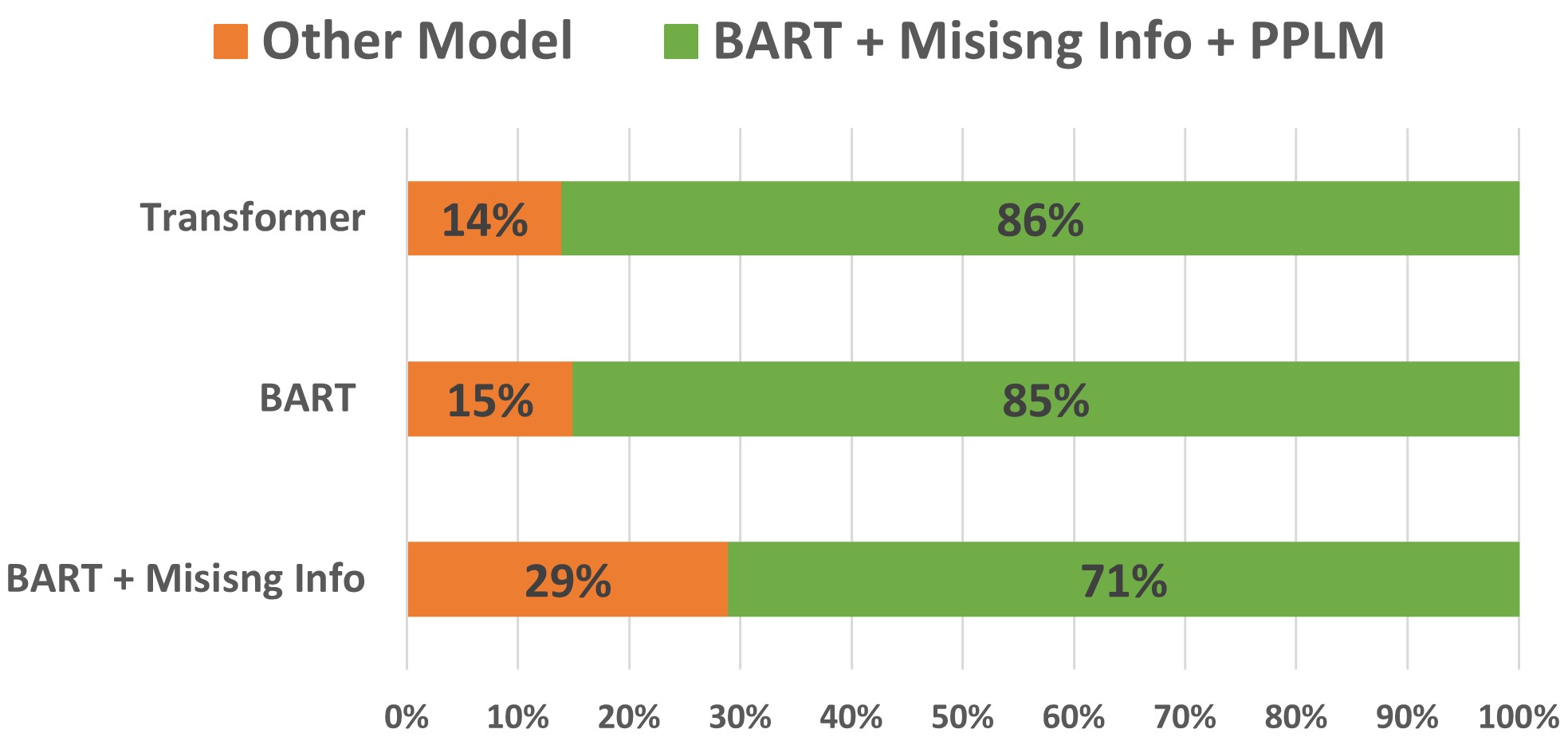}  
  \caption{}
  \label{fig:ubuntu-usefulness-compare-results}
\end{subfigure}
\caption{
Results of a pairwise comparison (on usefulness criteria) between our model and baseline generated question on (a) 300 randomly sampled product descriptions from 
the
Amazon test set, (b) 150 randomly sampled dialogs from 
the
Ubuntu test set as judged by humans.
}
\label{fig:fig}
\vspace{-1em}
\end{figure*}

\paragraph{Amazon} 
\autoref{tab:amazon-human-eval-table} shows the human judgment results on model generations for 300 randomly sampled product descriptions from the Amazon test set. Under relevancy and fluency, all models score reasonably with our proposed model producing the most relevant and fluent questions. Under missing information, the BART model, fine-tuned on context instead of missing schema, has the lowest score. GAN-Utility outperforms BART but significantly lags behind BART+missinfo and BART+missinfo+PPLM reaffirming our finding from the automatic metric results that our idea of feeding missing schema to a learning model helps.

\naacl{We additionally observe that the human-written questions score lower than model-generated questions under `fluency' and `missing information' criteria, mirroring similar observations from \cite{rao2018learning, rao-daume-iii-2019-answer}. We believe the reason for this is that human-written questions often have typos or are 
written by non-native speakers (leading to lower fluency). Moreover, humans may miss out on reading full product descriptions causing them to ask about details that are already included in the description (leading to lower missing information scores).}

\autoref{fig:amazon-usefulness-compare-results} shows the results of pairwise comparison on the usefulness criteria. We find that our model wins over GAN-Utility by a significant margin with humans preferring our model-generated questions 77\% of the time. 
Our model also beats BART-baseline 66\% of the time further affirming the importance of using missing schema. Finally, our model beats BART+missinfo model 61\%
of the time
suggesting that the PPLM-based decoder that uses usefulness classifier is able to produce much more \textit{useful} questions (RQ3). The annotator agreement statistics are provided in appendix. 

\paragraph{Ubuntu}
\autoref{tab:ubuntu-human-eval-table} shows the results of human judgments on the model generations of 150 randomly sampled dialog contexts from 
the
Ubuntu test set. 
In terms of relevance,
we find that the transformer and 
BART baselines produce less relevant questions. With
the
addition of missing schema (i.e.,~BART+missinfo), the questions become more relevant and our proposed model 
obtains
the highest 
relevance
score. The reference 
obtains
slightly 
a
lower 
relevance
score which can 
possibly
be explained by the fact that humans sometimes digress from the topic. Under fluency, interestingly, the transformer and BART baselines 
obtain
high scores. With 
the
addition of missing schema, fluency 
decreases
and the score 
reduce
further 
with the PPLM model. We suspect that the usefulness classifier trained with 
a
negative sampling strategy (as opposed to human labelled data, as in Amazon) contributes to fluency issues.
Under missing information, all models 
perform well
which can be explained by the fact that in Ubuntu, the scope of missing information is much larger (since dialog is much more open-ended) than in Amazon. 

\autoref{fig:ubuntu-usefulness-compare-results} shows the results of pairwise comparison on usefulness criteria. We find that humans choose our model-generated questions 85\% of time when compared to either transformer or BART generated questions.
When compared to BART+missinfo, our model 
is selected
71\% of the time, further affirming the importance of using the PPLM-based decoder.


\begin{table*}[t]
    \centering
    \footnotesize
    \begin{tabular}{l|l}
    \hline
        \textbf{Amazon} & \\
        \hline
        \hline
         Category & Binoculars \& Scopes \\
         Title & Nikon 7239 Action 7x50 EX Extreme All-Terain Binocular \\
         Description & The Monarch ATB 42mm with dielectric high-reflective Multilayer Prism coating  binocular \\
         &  features brighter, sharper colors, crisp and drastically improved low-light performance. \\
         &  A new body style provides unparalleled strength and ruggedness in a package ... \\
         Missing Schema & \{mounting, \textbf{center focused}, (Nikon, works, obj), (Canon, works, obj), digital camera, \ldots\} \\
         \hline
         GAN-Utility & price? \\
         BART & How is the focus quality?\\
         BART+missinfo & Is it \textbf{center focused}?\\
         BART+missinfo+PPLM & Is it \textbf{center focused}, or do you have to focus each eye individually? \\
         \hline
         \textbf{Ubuntu} & \\
        \hline
        \hline
        Dialog history & User A: I'm having trouble installing nvidia drivers for my geforce 6200, \\
        & could anyone perhaps assist? \\
        & User B: i use the drivers from the website, much better \\
        & User A: which drivers? from the website? \\
        & User B: I used add/remove software from the menu to install nvidia proprietary drivers\\
        Missing schema & \{(driver, update, nsubj), (new version, install, nsubj), (machine, \textbf{reboot}, nsubj), ...\}\\
        \hline
        Transformer & Did you try booting your machine?\\
        BART & where did you download them from?\\
        BART+missinfo & Can you tell the output after you install them?\\
        BART+missinfo+PPLM & Can you try \textbf{rebooting} from the start and removing the software after installation?\\
        \hline
    \end{tabular}
    \caption{Model generations for an example product from Amazon and an example dialog context from Ubuntu.}
    \label{tab:amazon-model-outputs}
\end{table*}

\begin{table}[h]
\centering
\footnotesize
\begin{tabular}
{l c c }
Model &  Amazon & Ubuntu  \\
\hline
\hline
Retrieval & 10.5 & 6.78 \\
GAN-Utility & 73.4 & -- \\
Transformer & 57.2 & 45.7\\
BART & 60.3 &  56.9\\
\hspace{0.6em} + missinfo & 97.3 & 89.2\\
\hspace{0.6em} + missinfo + PPLM & \bf 98.3 & \bf 90.1\\
Reference & 99.7 & 93.7\\
\hline
\end{tabular}
\caption{\label{tab:grounding-table} Missing information overlap (in \%) between missing schema and output generations}
\vspace{-1em}
\end{table}

\section{Analysis} \label{sec:analysis}

\paragraph{Robustness to input information} We analyze how a model is robust toward
the
amount of information present. To measure the amount of information, we look for context length (description length for Amazon, dialog context length for Ubuntu) and the size of global schema since these two directly control how much knowledge regarding potential missing information is available to the model.
We measure the difference in BLEU score between two groups of data samples where context length/size of global schema is either high or low.
\autoref{fig:bleu-diff} shows that our model is the least variant toward the information available hence more robust for
the
Amazon dataset.\footnote{Ubuntu follows similar trends; figure in appendix.} 

Owing to our modular approach for estimating missing information, we seek to analyze 
whether
a question is really asking about 
missing information in an automatic fashion. This also allows us to explain the reasoning behind a particular generation as we are able to trace back to the particular missing information that is used to generate the question.
We
run a YAKE extractor on the generated questions to obtain key-phrases. We calculate the ratio between the number of key-phrases in the output that belong to the original missing schema and 
the
total number of key-phrases present in the output. \autoref{tab:grounding-table} shows that when we use our framework of estimating missing 
information
coupled with BART, both models achieve very high missing information overlap, 
thus suggesting that we can obtain the reasoning behind a generated question reliably by tracing the missing information overlap, as shown in \autoref{tab:amazon-model-outputs}.

\paragraph{Question length} \naacl{We also observe in \autoref{tab:amazon-model-outputs} that baseline models tend to generate short and generic questions as compared to our model that often chooses longer schema key-phrases (e.g.~bigrams) to generate a more specific question. We further looked into annotated (for usefulness) questions from the Amazon dataset and we observed that 70\% of questions that were annotated as useful are longer than not-useful questions. The average length of gold useful questions is 10.76 words and 8.21 for not-useful questions. The average length of generated questions for BART, BART+MissInfo and BART+MissInfo+PPLM (ours) are 5.6, 6.2, 12.3 respectively. We also find a similar trend in the Ubuntu dataset as well.}

\paragraph{Dynamic expansion of global schema}
\naacl{We anticipate that even if we build the global schema from the available offline dataset, it is possible that new entries may appear in a real application. We investigate how our framework responds to the dynamic expansion of global schema.
We simulate a scenario where we extend the ``Laptop Accessories'' category in the Amazon dataset, with 100 new products (those that appeared on Amazon.com after the latest entry in the dataset). We obtain key-phrases from their product descriptions and include them in the global schema for the category which amounts to a 21\% change in the existing global schema. For 50 random products in the test set from the same category, we found that in 28 out of 50 cases (56\%), the model picked a new schema element that is added later. This indicates that our framework is capable of supporting dynamic changes in the global schema and reflecting them in subsequent generations without retraining from scratch.
}

\begin{figure}[t!]
  \centering
  \includegraphics[trim=0 0 0 0, width=\linewidth]{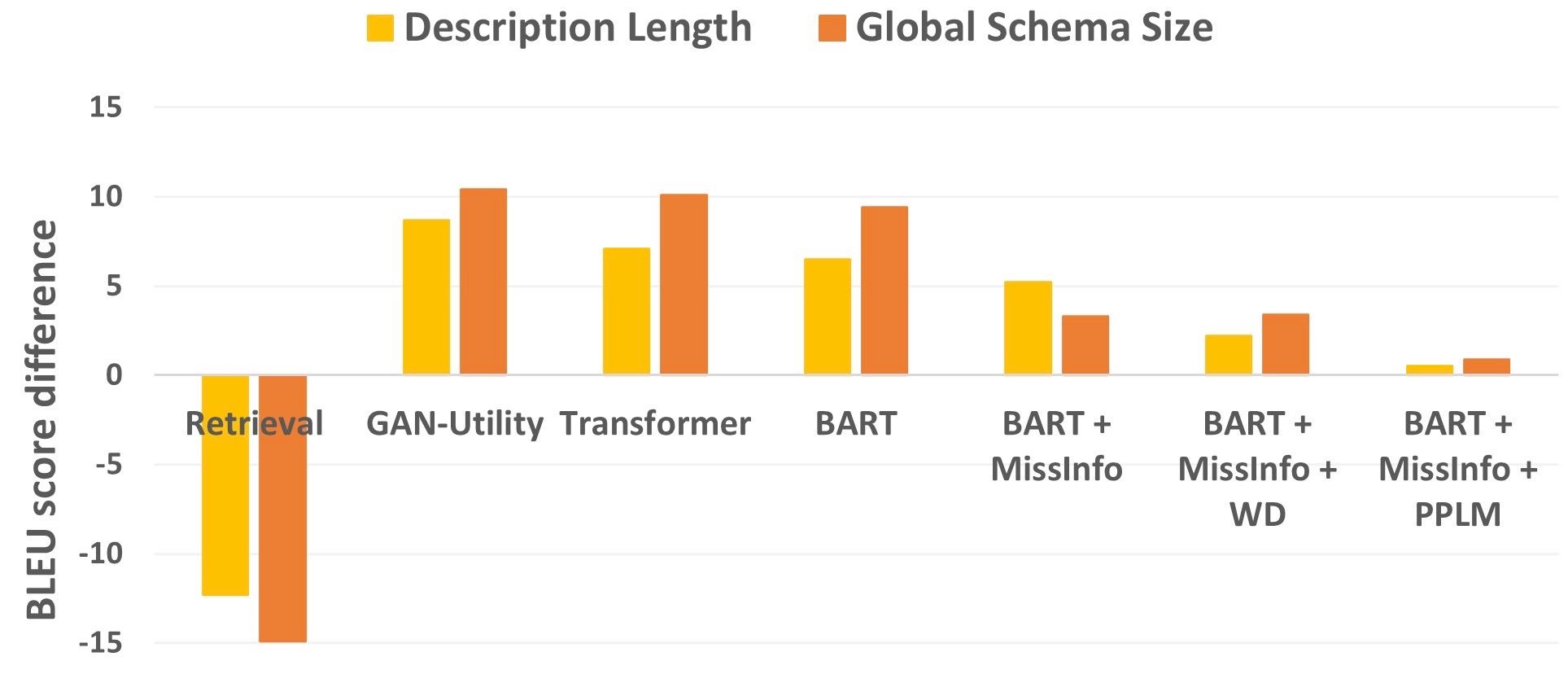} 
  \label{fig:amazon-bleu-diff}
\caption{\small
Average BLEU score difference between classes having longer ($>$ 200 (median) words) and shorter descriptions; larger ($>$ 200 (median) key-phrases) and shorter global schema for 
the
Amazon dataset. Lower differences indicate more invariance toward
the
available
information.
}
\label{fig:bleu-diff}
\vspace{-1em}
\end{figure}

\section{Related Work}


Most previous work on question generation focused on generating reading comprehension style questions i.e.,~questions that ask about information present in a given text \cite{DBLP:conf/emnlp/DuanTCZ17, DBLP:journals/corr/abs-1909-06356}. Later, \citet{rao2018learning, rao-daume-iii-2019-answer} introduced the task of clarification question generation in order to ask questions about missing information in a
given context. ClarQ \cite{DBLP:conf/acl/KumarB20} entails clarification questions in a question answering setup. However, unlike our work, these works still suffer from estimating the most useful missing information. 

Recent works on conversational question answering also focused on the aspect of question generation or retrieval \cite{DBLP:conf/emnlp/ChoiHIYYCLZ18, DBLP:conf/sigir/AliannejadiZCC19}.  
\citet{DBLP:conf/emnlp/0003ZM20} especially focused on generating information-seeking 
questions while 
\citet{DBLP:conf/emnlp/MajumderLNM20} proposed a question generation task in free-form interview-style conversations. 
In this work, in addition to improving clarification question generation in a community-QA dataset, we are 
the
first to explore a goal-oriented dialog scenario as well.

Representing context and associated global information in a structure format has been shown to improve performance in generation task \cite{DBLP:conf/iclr/DasMYTM19, DBLP:conf/acl/SubramanianWYZT18, khashabi2017learning} in general and summarization \cite{DBLP:conf/emnlp/FanGBB19} and story-generation \cite{DBLP:conf/aaai/YaoPWK0Y19} in particular. 
We also derive inspiration from recent works on information extraction from free-form text \cite{DBLP:journals/corr/abs-1904-08524, DBLP:journals/corr/StanovskyFDG16} and develop a novel framework to estimate missing information from available natural text contexts. 

Finally, for question generation, we use BART \cite{lewis2019bart}, that is state-of-the-art for many generation tasks such as summarization, dialog generation etc. Furthermore, inspired from recent works that use controlled language generation during decoding \cite{ghazvininejad2017hafez, DBLP:conf/acl/ChoiBGHBF18}, we use Plug-and-Play-Language-Model \cite{dathathri2019plug} to tune generations during decoding. While similar approaches for controllable generation \cite{DBLP:journals/corr/abs-1909-05858, DBLP:conf/naacl/SeeRKW19} have been proposed, we extend such efforts to enhance the usefulness of the generated clarification questions.

\section{Conclusion}
We propose a model for generating useful clarification questions based on the idea that missing information in a context can be identified by taking a difference between the global and the local view. We show how we can fine-tune a large-scale pretrained model such as BART on such differences to generate questions about missing information. Further, we show how we can tune these generations to make them more useful using PPLM with a usefulness classifier as its attribute model.
\naacl{Thorough analyses reveal that our framework works across domains, shows robustness towards information availability, and responds to the dynamic change in global knowledge.} 
Although we experiment only with Amazon and Ubuntu datasets, our idea is generalizable to scenarios where it is valuable to identify missing information such as conversational recommendation, or eliciting user preferences in a chit-chat, 
among others.

\paragraph{Acknowledgements} \naacl{We thank everyone in the Natural Language Processing Group at Microsoft Research, Redmond, with special mention to Yizhe Zhang, Bill Dolan, Chris Brockett, and Matthew Richardson for their critical review of this work. We also thank anonymous reviewers for providing valuable feedback. In addition to this, we want to acknowledge human annotators from Amazon Mechanical Turk for data annotation and human evaluation of our systems. BPM is partly supported by a Qualcomm Innovation Fellowship and NSF Award \#1750063. Findings and observations are of the authors only and do not necessarily reflect the views of the funding agencies.}

\section{Broader Impact}
We do not foresee any immediate ethical concerns
since we assume that 
our work
will be restricted in domain as compared to 
free-form language generation. We still cautiously 
advise
any developer who 
wishes
to 
extend
our system for their own use-case (beyond e-commerce, goal-oriented conversations) 
to
be careful about curating 
a global pool of knowledge 
for data involving sensitive user information.
Finally, since we are finetuning a pretrained generative model, 
we inherit the general
risk of generating biased or toxic language, which should be carefully filtered.
In general, 
we expect 
users to 
benefit
from
our system by 
reducing
ambiguity (when information is presented in a terse fashion, e.g.~in a conversation) and improving contextual understanding to enable them to take more informed actions (e.g.~making a purchase).



\bibliography{anthology,main}
\bibliographystyle{acl_natbib}

\appendix

\section{Setup and Data}
\paragraph{Schema}
While creating the global schema, we use word embedding\footnote{We train GLoVE embeddings separately on Amazon and Ubuntu} based similarity to perform hierarchical clustering of key-phrases\footnote{For triples, we use only their key-phrase to define similarity.} and group together key-phrases that have cosine similarity greater than a threshold, a hyperparameter set to $0.6$. Finally, we order all key-phrase clusters by their frequencies and retain only the top 60\% thus removing low-frequency schema elements. 

While creating the missing schema, we do the match based on semantic similarity of key-phrases (even for a tuple we only look at key-phrase similarity) and we consider two key-phrases to be matched if the cosine similarity is above a threshold, that we set as $0.8$ since we want to match only highly similar key-phrases. 

\paragraph{GloVe embeddings on Amazon and Ubuntu datasets}
We train 200 dimensional GLoVE embeddings on the vocabulary of both Amazon and Ubuntu dataset separately. We set a vocabulary frequency threshold at 50, i.e. we only obtain embeddings for words that appears at least 50 times in the whole corpus.

\paragraph{Datasets}
Downloadable links to each datasets are provided here: Amazon\footnote{\url{https://nijianmo.github.io/amazon/index.html}}, Ubuntu\footnote{\url{https://github.com/rkadlec/ubuntu-ranking-dataset-creator}}.

\paragraph{Collecting human annotations for usefulness scores}
For the Amazon dataset, \citet{rao-daume-iii-2019-answer} define the \textit{usefulness} of a question as the degree to which the answer provided by the question would be useful to potential buyers or current users of the product. We use the annotation scheme defined in \citet{rao-daume-iii-2019-answer} to annotate a set of 5000 questions from the amazon dataset.\footnote{We use the Amazon Mechanical Turk platform.} We show annotators product details (title, category, and description) and a question asked about that product and ask them to give it a usefulness score between 0 to 5.\footnote{Refer \citet{rao-daume-iii-2019-answer} for an exact description of each score.} Each question was annotated by three annotators. We average the three scores to get a single usefulness score per question. We use the YAKE extractor to extract the schema elements for each question and assign the usefulness score of the question to each of its schema elements. 

Since our aim is to assign a usefulness score to each missing element of each product in our dataset, we train a usefulness classifier on the manually annotated schema elements. Although our usefulness score is a real value between 0 and 5, we find that training a regression model gives us poor performance. Hence we convert the real value into a binary value by threshold at 3 (i.e. values below 3 are assigned label 0 and values above 3 are assigned label 1).

\paragraph{Usefulness classification with negative sampling}
Collecting usefulness annotation on questions, as we do for the Amazon dataset, can be expensive and may not always be possible in different scenarios. Therefore, for the Ubuntu dataset, we experiment with a classifier where instead of using human annotations are true labels, we use a negative sampling strategy. Specifically, we assume that all (context, question) pairs in the Ubuntu dataset can be labelled 1 and any (context, random question) can be labelled 0. We sample a set of 2500 questions from the Ubuntu dataset and them label 1 and sample an equivalent number of negative samples and assign them label 0.

\section{Training}

\paragraph{BART and PPLM}
For question generation model, we use BART-base (6 encoder layers, 6 decoder layers, 117M parameters) \footnote{\url{https://huggingface.co/transformers/model_doc/bart.html}}). For PPLM usefulness classifier, we use a bag-of-word model, that uses the pretrained subword embedding layers from BART-base model. We average the subword embeddings to obtain a sentence representation and a usefulness score is predicted via a linear layer projection with softmax. We use the the PPLM code from official repository\footnote{\url{https://github.com/uber-research/PPLM}}.

Each BART variant converged in 3 epochs on an average with batch size 4 in a TITAN X (Pascal) GPU that took 12 hours in total. While training, we only observe perplexity on the validation set to employ an early-stopping criteria.

\paragraph{Usefulness Classifier for BART+MissInfo+WD}
We train an SVM (support vector machines) classifier on this data. We use word emebddings as our features by training a 200 dimensional GLoVE model trained on individual dataset. We average the word embeddings of all words in a schema element and use it as a feature. We obtain an F1-score of 80.6\% on a held out test set.\footnote{In comparison, humans get an F1-score of 82.7\% in Amazon dataset} We use this classifier to predict a usefulness score for each missing schema element of each instances from a class for each dataset, which was required for the BART+MissInfo+WD model.

\section{More Experimental Analysis}
We additionally report Krippendorff’s alpha, a measure of annotator agreement for our human evaluation, on Amazon dataset.They are : for fluency 0.408, for relevancy 0.177, for missinginfo 0.226, and for usefulness 0.0948. For usefulness, we observe, if the systems are more distinct (GAN-Utility vs BART+missinfo+PPLM), then the agreement is higher i.e. 0.163. For missinginfo, again, 3-way gives higher agreement (0.434), and a probable cause would be that more annotations are going into the undecided category.

Additionally, \autoref{fig:bleu-diff} shows the BLEU difference across different data samples (based on context length and global schema size) that follow a similar trend to Amazon. \autoref{tab:amazon-model-outputs} shows generations from all the models, with a case the our best model trades off with missing information to improve the usefulness.

\begin{table*}[t]
    \centering
    \footnotesize
    \begin{tabular}{l|l}
    \hline
        \textbf{Amazon} & \\
        \hline
        \hline
         Category &  Bookshelf Speakers \\
         Title & Yamaha NS-6490 3-Way Bookshelf Speakers Finish (Pair) Black\\
         Description & Upgrade your current 5.1 home theater to a 7.1-Channel surround sound system by                    adding a  \\
                     & pair of Yamaha NS-6490 bookshelf speakers. This speaker was designed for both \\
                     & professional \& home entertainment enthusiasts with the capability to deliver a full, clear,... \\
         Missing Schema & \{\textbf{speaker wire}, \textbf{mounting}, (amplifier, tune, nsubj), wireless, bass, (iPhone, connect, obj), ...\} \\ 
         \hline
         GAN-Utility & are these speakers compatible with a yamaha satellite speakers? \\
         BART & What are the dimensions? \\
         BART+missinfo & Do the speakers come with \textbf{speaker wire}?\\
         BART+missinfo+PPLM & What kind \textbf{mounting} does this speaker use?\\
        \hline
        \hline
        Category &  Camera \& Photo \\
         Title & Porta Trace 10 x 12-inches Stainless Steel Frame Lightbox with Two 5000K Lamps\\
         Description & Gagne Porta-Trace light boxes virtually eliminate the hot spots found in competitive lightbox\\ & units. Redesigned frame and reflector combine with the thick Plexiglas top to provide uniform and even \\
         & lighting over the entire durable, stable viewing surface.\\
         & Durable Stainless Steel frame will maintain its attractive appearance for years...
         \\
         Missing Schema & \{camera, \textbf{battery powered}, flash, wireless, canon, nikon, ...\} \\ 
         \hline
         GAN-Utility & will this work with a canon rebel? \\
         BART & Does it come with the bulbs? \\
         BART+missinfo & Is it \textbf{battery powered}?\\
         BART+missinfo+PPLM & Can I replace the bulbs?\\
        \hline
    \end{tabular}
    \caption{Model generations for two examples product from Amazon. In the second example, our best model trades off with missing information to make the question more useful.}
    \label{tab:amazon-model-outputs}
\end{table*}

\begin{figure}[t]
  \centering
  \includegraphics[trim=0 0 0 0, width=0.95\linewidth]{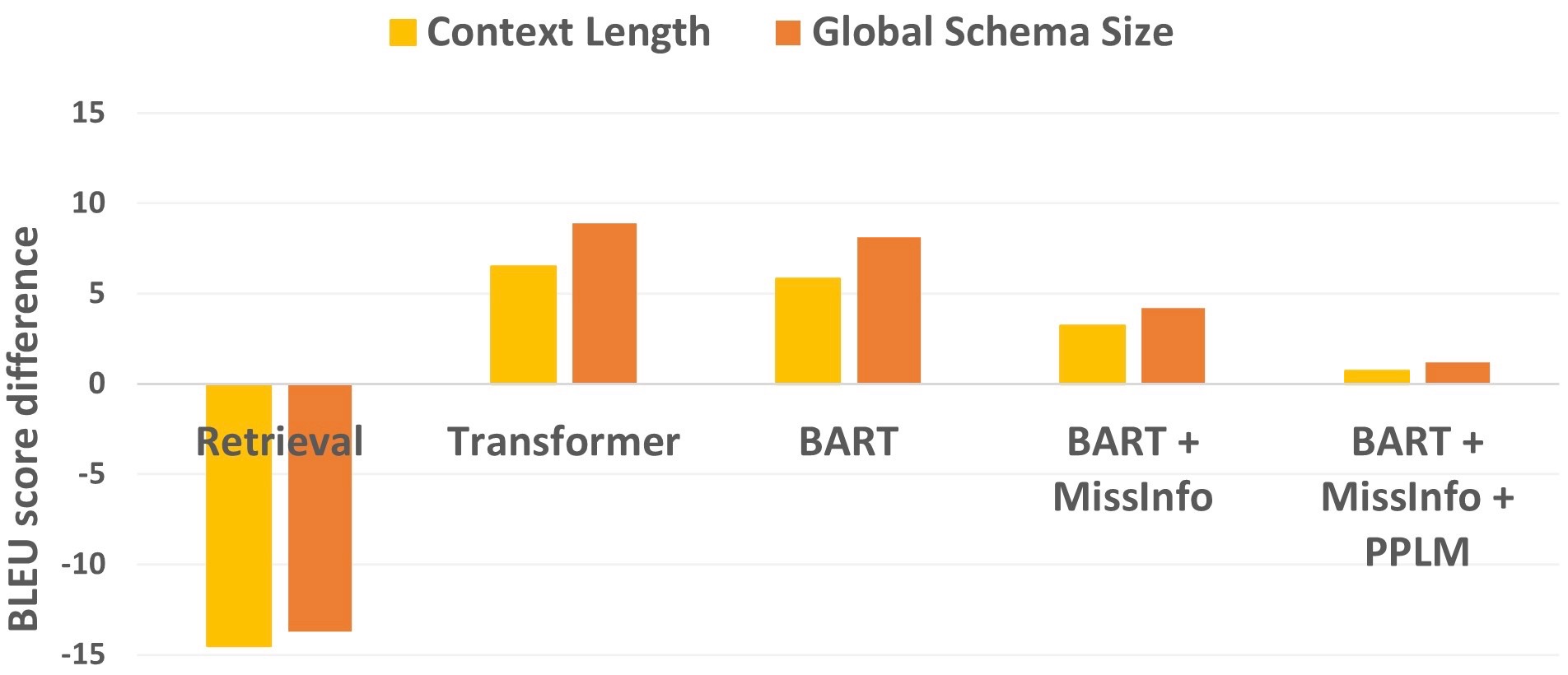} 
  \label{fig:ubuntu-bleu-diff}
\caption{
Average BLEU score difference between classes having longer (more than 200 (median) words) and shorter descriptions larger (more than 200 (median) key-phrases) and shorter global schema for Ubuntu dataset. Lower difference indicates more invariance towards information available.
}
\label{fig:bleu-diff}
\end{figure}




\end{document}